%% file: main.tex
\documentclass{article}

    \usepackage[preprint]{neurips_2019}

\usepackage[utf8]{inputenc} % allow utf-8 input
\usepackage[T1]{fontenc}    % use 8-bit T1 fonts
\usepackage{hyperref}       % hyperlinks
\usepackage{url}            % simple URL typesetting
\usepackage{booktabs}       % professional-quality tables
\usepackage{caption}

\DeclareCaptionLabelFormat{andfigure}{#1~#2  \&  \figurename~\thefigure}

\usepackage{tabularx}
\usepackage[export]{adjustbox}

\usepackage{multirow}
\usepackage{comment}
\usepackage{makecell}
\usepackage{amsfonts}       % blackboard math symbols
\usepackage{amsmath}
\usepackage{nicefrac}       % compact symbols for 1/2, etc.
\usepackage{microtype}      % microtypography
\usepackage{graphicx}      % images
\usepackage[english]{babel}
\usepackage[dvipsnames]{xcolor}
\usepackage[normalem]{ulem}
\newif{\ifhidecomments}

\hidecommentstrue
\ifhidecomments
    \newcommand{\jdcomment}[1]{}
    \newcommand{\vscomment}[1]{} 
    \newcommand{\resolved}[1]{}
\else
    \newcommand{\jdcomment}[1]{\textcolor{teal}{[#1 ({\bf Organisers})]}} 
    \newcommand{\vscomment}[1]{\textcolor{red}{[#1 ({\bf Varun})]}} 
    \newcommand{\resolved}[1]{} 
\fi

\newcommand{\tableaboveskip}{7pt}
\newcommand{\tablebelowskip}{0pt}
\newcommand{\figureaboveskip}{0pt}
\newcommand{\figurebelowskip}{7pt}

\title{Reproducibility Report \\Rigging the Lottery: Making All Tickets Winners}

\author{%
  Varun Sundar \\
  University of Wisconsin Madison\\
  \texttt{vsundar4@wisc.edu} \\
   \And
   Rajat Vadiraj Dwaraknath \\
   Stanford University \\
   \texttt{rajatvd@stanford.edu} \\
}

\begin{document}

\maketitle

\input{sections/front_page}
\input{sections/introduction}
\input{sections/scope}
\input{sections/methodology}

\input{sections/experiments}
\input{sections/results}

\input{sections/results_beyond}
\input{sections/discussion}

\medskip

% \bibliography{ref}

\input{main.bbl}
\end{document}

\maketitle

\input{supplementary_sections/architecture_details}
\input{supplementary_sections/flop_counting}
\input{supplementary_sections/hyperparameter_tuning}
\input{supplementary_sections/structured_sparsity}

\medskip

% \bibliography{ref}

\input{supplementary.bbl}

%% file: sections/front_page.tex
\section*{\centering Reproducibility Summary}

% \textit{Template and style guide to \href{https://paperswithcode.com/rc2020}{ML Reproducibility Challenge 2020}. The following section of Reproducibility Summary is \textbf{mandatory}. This summary \textbf{must fit} in the first page, no exception will be allowed. When submitting your report in OpenReview, copy the entire summary and paste it in the abstract input field, where the sections must be separated with a blank line.}

\subsection*{Scope of Reproducibility}

% State the main claim\jdcomment{(s)} of the original paper you are trying to reproduce. We recommend picking the central claim\jdcomment{(s)} of the paper. \jdcomment{For those that are familiar, this can be framed as scientific hypotheses (i.e. falsifiable, clear how evidence will support the hypothesis, etc.).}

% For a fixed parameter count and compute budget, the proposed algorithm, \textit{RigL} directly trains sparse networks with constant Floating Point Operations (FLOPs) throughout training, while matching the performance of existing dense-to-sparse training techniques. The technique obtains state-of-the-art performance on a variety of tasks, including image classification and character-level language-modelling.

For a fixed parameter count and compute budget, the proposed algorithm (\textit{RigL}) claims to directly train sparse networks that match or exceed the performance of existing dense-to-sparse training techniques (such as pruning). \textit{RigL} does so while requiring constant Floating Point Operations (FLOPs) throughout training. The technique obtains state-of-the-art performance on a variety of tasks, including image classification and character-level language-modelling.

\subsection*{Methodology}

% Briefly describe what you did and which resources you used. Did you use author's code? Did you re-implement parts of the pipeline? How much time did it take to produce the results? What hardware you were using and how long did it take (e.g. GPU hours) to train/evaluate? 

We implement \textit{RigL} from scratch in Pytorch using boolean masks to simulate unstructured sparsity. We rely on the description provided in the original paper, and referred to the authors' code for only specific implementation detail such as handling overflow in ERK initialization. We evaluate sparse training using \textit{RigL} for WideResNet-22-2 on CIFAR-10 and ResNet-50 on CIFAR-100, requiring 2 hours and 6 hours respectively per training run on a GTX 1080 GPU.

\subsection*{Results}

% Start with your overall conclusion - where \jdcomment{did your study reproduce results from the original paper, and where did your results differ?} \sout{was your study successful and where not successful.} Be specific and use precise language, e.g. "we reproduced the accuracy to within 1\% of reported value, which supports the paper's conclusion that it outperforms the baselines". Getting exactly the same number is in most cases infeasible, so you'll need to use your judgement to decide if your results support the original claim of the paper.

We reproduce \textit{RigL}'s performance on CIFAR-10 within 0.1\% of the reported value. On both CIFAR-10/100, the central claim holds---given a fixed training budget, \textit{RigL} surpasses existing dynamic-sparse training methods over a range of target sparsities. By training longer, the performance can match or exceed iterative pruning, while consuming constant FLOPs throughout training. We also show that there is little benefit in tuning \textit{RigL}'s hyper-parameters for every sparsity, initialization pair---the reference choice of hyperparameters is often close to optimal performance. 

Going beyond the original paper, we find that the optimal initialization scheme depends on the training constraint. While the Erdos-Renyi-Kernel distribution outperforms Random distribution for a fixed parameter count, for a fixed FLOP count, the latter performs better. Finally, redistributing layer-wise sparsity while training can bridge the performance gap between the two initialization schemes, but increases computational cost.

\subsection*{What was easy}

% Describe which parts of your reproduction study were easy. For example, was it easy to run the author's code, or easy to re-implement their method based on the description in the paper? The goal of this section is to summarize to a reader which parts of the original paper they could easily apply to their problem.

The authors provide code for most of the experiments presented in the paper. The code was easy to run and allowed us to verify the correctness of our re-implementation. The paper also provided a thorough and clear description of the proposed algorithm without any obvious errors or confusing exposition. 

\subsection*{What was difficult}

% Describe which parts of your reproduction study were difficult or took much more time than you expected. Perhaps the data was not available and you couldn't verify some experiments, or the author's code was broken and had to be debugged first. Or, perhaps some experiments just take too much time/resources to run and you couldn't verify them. The purpose of this section is to indicate to the reader which parts of the original paper are either difficult to re-use, or require a significant amount of work and resources to verify. 

Tuning hyperparameters involved multiple random seeds and took longer than anticipated. Verifying the correctness of a few baselines was tricky and required ensuring that the optimizer's gradient (or momentum) buffers were sparse (or dense) as specified by the algorithm. Compute limits restricted us from evaluating on larger datasets such as Imagenet.

\subsection*{Communication with original authors}

% Briefly describe how much contact you had with the original authors (if any).

We had responsive communication with the original authors, which helped clarify a few implementation and evaluation details, particularly regarding the FLOP counting procedure.

%% file: sections/introduction.tex
\section{Introduction}

\jdcomment{A  few  sentences  placing  the  work  in  context. Limit it to a few paragraphs at most; your report is on reproducing a piece of work, you don’t have to motivate that work.}

Sparse neural networks are a promising alternative to conventional dense networks---having comparatively greater parameter efficiency and lesser floating-point operations (FLOPs) (\citet{han2016eie,ashbyexploiting,Srinivas_2017_CVPR_Workshops}). Unfortunately, present techniques to produce sparse networks of commensurate accuracy involve multiple cycles of training dense networks and subsequent pruning. Consequently, such techniques offer no advantage over training dense networks, either computationally or memory-wise. 

In the paper \citet{rigl}, the authors propose \textit{RigL}, an algorithm for training sparse networks from scratch. The proposed method outperforms both prior art in training sparse networks, as well as existing dense-to-sparse training algorithms. By utilising dense gradients only during connectivity updates and avoiding any global sparsity redistribution, \textit{RigL} can maintain a fixed computational cost and parameter count throughout training.

As a part of the ML Reproducibility Challenge, we replicate \textit{RigL} from scratch and investigate if dynamic-sparse training confers significant practical benefits compared to existing sparsifying techniques.

% for a fixed training budget. 

%% file: sections/scope.tex
\section{Scope of reproducibility}
\label{sec:claims}

\jdcomment{Explain the claims from the original paper you picked for the reproduction study and briefly motivate your choice. Try to summarize each claim  in 1-2 sentences, e.g. "The introduced activation function X outperforms a similar activation function Y on tasks Z,V,W". Make the scope as specific as possible. It should be something that can be supported or rejected by your data. For example, this scope is too broad and lacks precise outcome: "Contextual embedding models have shown strong performance on a number of tasks across NLP. We will run experiments evaluating two types of contextual embedding models on datasets X, Y, and Z." This scope is better because it's more specific and has an outcome that can be either supported or rejected based on your results: "Finetuning pretrained BERT on SST-2 will have higher accuracy than an LSTM trained with GloVe embeddings."}

In order to verify the central claims presented in the paper we focus on the following target questions:

\begin{itemize}
    \item Does \textit{RigL} outperform existing sparse-to-sparse training techniques---such as SET (\citet{Mocanu2018SET}) and SNFS (\citet{dettmers2020sparse})---and match the accuracy of dense-to-sparse training methods such as iterative pruning (\citet{to_prune_or_not})?
    
    \item \textit{RigL} requires two additional hyperparameters to tune. We investigate the sensitivity of final performance to these hyperparameters across a variety of target sparsities (Section \ref{hyperparameter-tuning}).
    
    \item How does the choice of sparsity initialization affect the final performance for a fixed parameter count and a fixed training budget (Section \ref{effect-sparsity-distribution})?
    
    \item Does redistributing layer-wise sparsity during connection updates (\citet{dettmers2020sparse}) improve \textit{RigL}'s performance? Can the final layer-wise distribution serve as a good sparsity initialization scheme (Section \ref{effect-redistribution})? 

\end{itemize}

\jdcomment{Each experiment in Section~\ref{sec:results} will support (at least) one of these claims, so a reader of your report should be able to understand the \emph{arguments} (claims) and, separately, the \emph{evidence} that supports them.}
%\jdcomment{To organizers: I asked my students to connect the main claims and the experiments that supported them. For example, in this list above they could have ``Claim 1, which is supported by Experiment 1 in Figure 1.'' The benefit was that this caused the students to think about what their experiments were showing (as opposed to blindly rerunning each experiment and not considering how it fit into the overall story), but honestly it seemed hard for the students to understand what I was asking for.}

%% file: sections/methodology.tex
\section{Methodology}

\jdcomment{Explain your approach - did you use the author's code, or did you aim to re-implement the approach from the description in the paper? Summarize the resources (code, documentation, GPUs) that you used.}

The authors provide publicly accessible code\footnote{\href{https://github.com/google-research/rigl}{https://github.com/google-research/rigl}} written in Tensorflow (\citet{abadi2016tensorflow}). To gain a better understanding of various implementation aspects, we opt to replicate \textit{RigL} in Pytorch (\citet{Pytorch}). Our implementation extends the open-source code\footnote{\href{https://github.com/TimDettmers/sparse_learning}{https://github.com/TimDettmers/sparse\_learning}} of \citet{dettmers2020sparse} which uses a boolean mask to simulate unstructured sparsity. Our source code is publicly accessible on Github\footnote{\href{https://github.com/varun19299/rigl-reproducibility}{https://github.com/varun19299/rigl-reproducibility}} with training plots available on WandB\footnote{\href{https://wandb.ai/ml-reprod-2020}{https://wandb.ai/ml-reprod-2020}} (\citet{wandb}).

\paragraph{Mask Initialization} For a network with $L$ layers and total parameters $N$, we associate each layer with a random boolean mask of sparsity $s_l, \; l \in [L]$. The overall sparsity of the network is given by $S=\frac{\sum_l s_l N_l}{N}$, where $N_l$ is the parameter count of layer $l$. Sparsities $s_l$ are determined by the one of the following mask initialization strategies:

\begin{itemize}
    \item \textbf{Uniform:} Each layer has the same sparsity, i.e., $s_l = S \; \forall l$. Similar to the original authors, we keep the first layer dense in this initialization.
    
    \item \textbf{Erdos-Renyi (ER):} Following \citet{Mocanu2018SET}, we set $s_l \propto \left(1 - \frac{C_\text{in} + C_\text{out}}{C_\text{in} \times C_\text{out}} \right)$, where $C_\text{in}, C_\text{out}$ are the in and out channels for a convolutional layer and input and output dimensions for a fully-connected layer. 
    
    \item \textbf{Erdos-Renyi-Kernel (ERK):} Modifies the sparsity rule of convolutional layers in ER initialization to include kernel height and width, i.e., $s_l \propto \left(1 - \frac{C_\text{in} + C_\text{out} + w + h}{C_\text{in} \times C_\text{out} \times w \times h} \right)$, for a convolutional layer with $C_\text{in} \times C_\text{out} \times w \times h$ parameters. 
\end{itemize}

We do not sparsify either bias or normalization layers, since these have a negligible effect on total parameter count.

\paragraph{Mask Updates} Every $\Delta T$ training steps, certain connections are discarded, and an equal number are grown. Unlike SNFS (\citet{dettmers2020sparse}), there is no redistribution of layer-wise sparsity, resulting in constant FLOPs throughout training.

\paragraph{Pruning Strategy} Similar to SET and SNFS, \textit{RigL} prunes $f$ fraction of smallest magnitude weights in each layer.  As detailed below, the fraction $f$ is decayed across mask update steps, by cosine annealing:

\begin{equation}
        f(t) = \frac{\alpha}{2} \left(1 + \cos \left(\frac{t\pi}{T_\text{end}} \right) \right)
\end{equation}

where, $\alpha$ is the initial pruning rate and $T_\text{end}$ is the training step after which mask updates are ceased.

\paragraph{Growth Strategy} \textit{RigL}'s novelty lies in how connections are grown: during every mask update, $k$ connections having the largest absolute gradients among current inactive weights (previously zero + pruned) are activated. Here, $k$ is chosen to be the number of connections dropped in the prune step. This requires access to dense gradients at each mask update step. Since gradients are not accumulated (unlike SNFS), \textit{RigL} does not require access to dense gradients at \textit{every} step. Following the paper, we initialize newly activated weights to zero.

%% file: sections/experiments.tex
\section{Experimental Settings}

\subsection{Model descriptions}
\jdcomment{Include a clear description of the mathematical setting, algorithm, and/or model. For each model or algorithm, be sure to include 1) the number of parameters (this is similar to space complexity), and 2) some measure of average runtime (this is similar to time complexity) on what hardware you have (for example, average time to predict labels for 100 instances from dataset X, on GPU Y, with batch size Z).}

For experiments on CIFAR-10 (\citet{Krizhevsky09learningmultiple}), we use a Wide Residual Network (\citet{Wide_ResNet_BMVC2016_87}) with depth 22 and width multiplier 2, abbreviated as WRN-22-2. For experiments on CIFAR-100 (\citet{Krizhevsky09learningmultiple}), we use a modified variant of ResNet-50 (\citet{He_2016_CVPR}), with the initial $7\times 7$ convolution replaced by two $3 \times 3$ convolutions (architecture details provided in the supplementary material). 

\subsection{Datasets and Training descriptions}
\jdcomment{For each dataset include 1) relevant statistics such as the number of examples and label distributions, 2) details of train/dev/test splits, 3) an explanation of any preprocessing done, and 4) a link to download the data (if available).}

We conduct our experiments on the CIFAR-10 and CIFAR-100 image classification datasets. For CIFAR-10, we use a train/val/test split of 45k/5k/10k samples. In comparison, the authors use no dedicated validation set, with 50k samples and 10k samples comprising the train set and test set, respectively. This causes a slight performance discrepancy between our reproduction and the metrics reported by the authors (dense baseline has a test accuracy of 93.4\% vs 94.1\% reported). However, our replication matches the paper's performance when 50k samples are used for the train set (Table \ref{tab:replication_verify}). We use a validation split of 10k samples for CIFAR-100 as well.

\input{tables/replication_verify}

On both datasets, we train models for 250 epochs each, optimized by SGD with momentum.  Our training pipeline uses standard data augmentation, which includes random flips and crops. When training on CIFAR-100, we additionally include a learning rate warmup for 2 epochs and label smoothening of 0.1 (\citet{goyal2017accurate}). We also initialize the last batch normalization layer (\citet{ioffe2015batch}) in each BottleNeck block to 0, following \citet{He_2019_CVPR}.

\subsection{Hyperparameters}
\jdcomment{Describe how the hyperparameter values were set. If there was a hyperparameter search done, be sure to include the range of hyperparameters searched over, the method used to search (e.g. manual search, random search, Bayesian optimization, etc.), and the best hyperparameters found. Include the number of total experiments (e.g. hyperparameter trials). You can also include all results from that search (not just the best-found results). Describe how you set the hyperparameters and what was the source for their value (e.g. paper, code or your guess).}

\textit{RigL} includes two additional hyperparameters ($\alpha, \Delta T$) in comparison to regular dense network training. In Sections \ref{cifar-10-results} and \ref{cifar-100-results}, we set $\alpha=0.3, \Delta T = 100$, based on the original paper. Optimizer specific hyperparameters---learning rate, learning rate schedule, and momentum---are also set according to the original paper. In Section \ref{hyperparameter-tuning}, we tune these hyperparameters with Optuna (\citet{optuna_2019}). We also examine whether indivdually tuning the learning rate for each sparsity value offers any significant benefit.

\subsection{Baseline implementations}
\jdcomment{Include a description of how the experiments were set up that's clear enough a reader could replicate the setup. 
Include a description of the specific measure used to evaluate the experiments (e.g. accuracy, precision@K, BLEU score, etc.). 
Provide a link to your code.} 

We compare \textit{RigL} against various baselines in our experiments: SET (\citet{Mocanu2018SET}), SNFS (\citet{dettmers2020sparse}), and Magnitude-based Iterative-pruning (\citet{to_prune_or_not}). We also compare against two weaker baselines, viz., \textit{Static Sparse} training and \textit{Small-Dense} networks. The latter has the same structure as the dense model but uses fewer channels in convolutional layers to lower parameter count. We implement iterative pruning with the pruning interval kept same as the masking interval for a fair comparison. 

\subsection{Computational requirements}
\jdcomment{Provide information on computational requirements for each of your experiments. For example, the total number of CPU/GPU hours and amount of memory used for each experiment (note: you'll have to record this as you run your experiments, so it's better to think about it ahead of time). Consider the perspective of a reader who wants to use the approach described in the paper -- list what they would find useful to understand what resources they would need.
\sout{Include a description of the hardware used, such as the GPU or CPU the experiments were run on.
You'll need to think about this ahead of time, and write your code in a way that captures this information so you can later add it to this section.}}

We run our experiments on a SLURM cluster node---equipped with 4 NVIDIA GTX1080 GPUs and a 32 core Intel CPU. Each experiment on CIFAR-10 and CIFAR-100 consumes about 1.6 GB and 7 GB of VRAM respectively and is run for 3 random seeds to capture performance variance. We require about 6 and 8 days of total compute time to produce all results, including hyper-parameter sweeps and extended experiments, on CIFAR-10 and CIFAR-100 respectively.

%% file: tables/replication_verify.tex
\begin{table}[t]
    \captionsetup{aboveskip=\tableaboveskip,belowskip=\tablebelowskip}
    \caption{\textbf{Test accuracy of reference and our implementations on CIFAR-10,} tabulated for three different sparsities. Note that the runs listed here do not use a separate validation set while training.}
    \label{tab:replication_verify}
    \centering

    \begin{tabular}{ c ccc ccc }
    \toprule
    \textbf{Method}& 
    \multicolumn{3}{c}{\textbf{Ours}} & \multicolumn{3}{c}{\textbf{Original}} \\
    \midrule
    {Dense} & 
    \multicolumn{3}{c}{94.6} & \multicolumn{3}{c}{94.1} \\
    \midrule
    {} & 
    \makecell{$1-s=0.1$}  & \makecell{$1-s=0.2$} & \makecell{$1-s=0.5$} &
    \makecell{$1-s=0.1$}  & \makecell{$1-s=0.2$} & \makecell{$1-s=0.5$} \\
    \cmidrule(lr){2-4} \cmidrule(lr){5-7}
    Static (ERK) & 
    91.6 & 93.2 & 94.3 &  
    91.6 & 92.9 & 94.2 \\
    
    Pruning & 
    93.2 & 93.6 & 94.3 & 
    93.3 & 93.5 & 94.1 \\
    
    RigL (ERK) & 
    93.2 & 93.8 & 94.4 & 
    93.1 & 93.8 & 94.3 \\
    \bottomrule

    \end{tabular}
\end{table}

%% file: sections/results.tex
\section{Results}
\label{sec:results}

Given a fixed training FLOP budget, \textit{RigL} surpasses existing dynamic sparse training methods over a range of target sparsities, on both CIFAR-10 and 100 (Sections \ref{cifar-10-results}, \ref{cifar-100-results}). By training longer, \textit{RigL} matches or marginally outperforms iterative pruning. However, unlike pruning, its FLOP consumption is constant throughout. This a prime reason for using sparse networks, and makes training larger networks feasible. Finally, as evaluated on CIFAR-10, the original authors' choice of hyper-parameters are close to optimal for multiple target sparsities and initialization schemes (Section \ref{hyperparameter-tuning}).

\jdcomment{Start with a high-level overview of your results. \sout{Does your work support the claims you listed in section 2?} Keep this section as factual and precise as possible, reserve your judgement and discussion points for the next "Discussion" section. 

\textbf{Results reproducing original paper}
For each experiment, say 1) which claim in Section~\ref{sec:claims} it supports, and 2) if it successfully reproduced the associated experiment in the original paper. \sout{how it relates to one of the claims and explain what your result is.} 
For example, an experiment training and evaluating a model on a dataset may support a claim that that model outperforms some baseline.
Logically group related results into sections.} 

\input{sections/results_cifar10}
\input{sections/results_cifar100}

\input{sections/hyperparameter_tuning}

%% file: sections/results_cifar10.tex
\subsection{WideResNet-22 on CIFAR-10}\label{cifar-10-results}

\input{tables/cifar10}
   
Results on the CIFAR-10 dataset are provided in Table \ref{tab:cifar10-main-results}. Tabulated metrics are averaged across 3 random seeds and reported with their standard deviation. All sparse networks use random initialization, unless indicated otherwise.

While SET improves over the performance of static sparse networks and small-dense networks, methods utilizing gradient information (SNFS, \textit{RigL}) obtain better test accuracies. SNFS can outperform \textit{RigL}, but requires a much larger training budget, since it (a) requires dense gradients at each training step, (b) redistributes layer-wise sparsity during mask updates. For all sparse methods, excluding SNFS, using ERK initialization improves performance, but with increased FLOP consumption. We calculate theoretical FLOP requirements in a manner similar to \citet{rigl} (exact details in the supplementary material). 

Figure \ref{fig:cifar10-main-results} contains test accuracies of select methods across two additional sparsity values: ($0.5, 0.95$). At lower sparsities (higher densities), \textit{RigL} matches the performance of the dense baseline. Performance further improves by training for longer durations. Particularly, training \textit{RigL} (ERK) twice as long at 90\% sparsity exceeds the performance of iterative pruning while requiring similar theoretical FLOPs. This validates the original authors' claim that \textit{RigL} (a sparse-to-sparse training method) outperforms pruning (a dense-to-sparse training method). 

\input{figs/cifar10_main}

%% file: tables/cifar10.tex
\begin{table}[t]
    \captionsetup{aboveskip=\tableaboveskip,belowskip=\tablebelowskip}
    \caption{\textbf{WideResNet-22-2 on CIFAR10}, tabulated for two density $(1-s)$ values. We group methods by their FLOP requirement and in each group, we mark the best accuracy in bold. Similar to \citet{rigl}, we assume that algorithms utilize sparsity during training. All results are obtained by methods implemented in our unified codebase.}
    \label{tab:cifar10-main-results}
    \centering
    
    \begin{tabular}{ c cc cc }
    \toprule
    \multirow{3}{*}{\textbf{Method}}& 
    \multicolumn{2}{c}{$\mathbf{1 - s=0.1}$} & \multicolumn{2}{c}{$\mathbf{1 - s=0.2}$} \\
    \cmidrule(lr){2-3} \cmidrule(lr){4-5}
    {} & 
    \makecell{Accuracy $\uparrow$ \\ (Test)}  & \makecell{FLOPs $\downarrow$  \\ (Train, Test)} &
    \makecell{Accuracy $\uparrow$ \\ (Test)}  & \makecell{FLOPs $\downarrow$  \\ (Train, Test)} \\
    \midrule
    Small Dense & 
    {89.0 $\pm$ 0.35} & {0.11x, 0.11x} & 
    {91.0 $\pm$ 0.07} & {0.20x, 0.20x} \\
    
    Static & 
    {89.1 $\pm$ 0.17} & {0.10x, 0.10x} & 
    {91.2 $\pm$ 0.16} & {0.20x,0.20x} \\

    SET &
    {91.3 $\pm$ 0.47} & {0.10x, 0.10x} & 
    \textbf{92.7 $\pm$ 0.28} & {0.20x, 0.20x} \\
    
    \textbf{RigL} &
    \textbf{91.7 $\pm$ 0.18} & {0.10x, 0.10x} &
    {92.6 $\pm$ 0.10} & {0.20x, 0.20x} \\
    \midrule
    
    SET (ERK)&
    {92.2 $\pm$ 0.04} & {0.17x, 0.17x} &
    {92.9 $\pm$ 0.16} & {0.35x, 0.35x} \\
    
    \textbf{RigL (ERK)} &
    \textbf{92.4 $\pm$ 0.06} & {0.17x, 0.17x} &
    \textbf{93.1 $\pm$ 0.09} & {0.35x, 0.35x} \\
    \midrule
    {Static\textsubscript{$2 \times$}} &
    {89.15 $\pm$ 0.17} & {0.20x, 0.10x} &
    {91.2 $\pm$ 0.16} & {0.40x, 0.20x} \\
    
    Lottery & 
    {90.4 $\pm$ 0.09} & {0.45x, 0.13x} & 
    {92.0 $\pm$ 0.31} & {0.68x,0.27x} \\
    
    {SET\textsubscript{$2 \times$}} &
    {83.3 $\pm$ 15.33} & {0.20x, 0.10x} &
    {93.0 $\pm$ 0.22} & {0.41x, 0.20x} \\
    
    SNFS & 
    {92.4 $\pm$ 0.43} & {0.51x, 0.27x} & 
    {92.7 $\pm$ 0.20} & {0.66x, 0.49x} \\ 
    
    SNFS (ERK)& 
    {92.2 $\pm$ 0.2} & {0.52x, 0.28x} & 
    {92.8 $\pm$ 0.07} & {0.66x, 0.49x} \\
    
    {SNFS\textsubscript{$2 \times$}} &
    {92.3 $\pm$ 0.33} & {1.02x, 0.27x} &
    {93.2 $\pm$ 0.14} & {1.32x, 0.98x} \\
    
    {RigL\textsubscript{$2 \times$}} &
    {92.3 $\pm$ 0.25} & {0.20x, 0.10x} &
    {93.0 $\pm$ 0.21} & {0.41x, 0.20x} \\
    
    % {RigL\textsubscript{$3 \times$}} &
    % {92.5 $\pm$ 0.11} & {0.30x, 0.10x} &
    % {93.2 $\pm$ 0.20} & {0.61x, 0.20x} \\
    
    {Pruning} & 
    {92.6 $\pm$ 0.08} & {0.32x,0.13x} & 
    {93.2 $\pm$ 0.27} & {0.41x,0.27x} \\ 
    
    \textbf{RigL\textsubscript{$2 \times$} (ERK)} &
    \textbf{92.7 $\pm$ 0.37} & {0.34x, 0.17x} &
    \textbf{93.3 $\pm$ 0.09} & {0.70x, 0.35x} \\
    \midrule
    
    \textbf{Dense Baseline} &
    \textbf{93.4 $\pm$ 0.07} & {9.45e8, 3.15e8} &
    \textbf{-} & {-} \\
    \bottomrule
    
    \end{tabular}
\end{table}

%% file: figs/cifar10_main.tex
\begin{figure}[!t]
    \centering
    \includegraphics[width=1\textwidth]{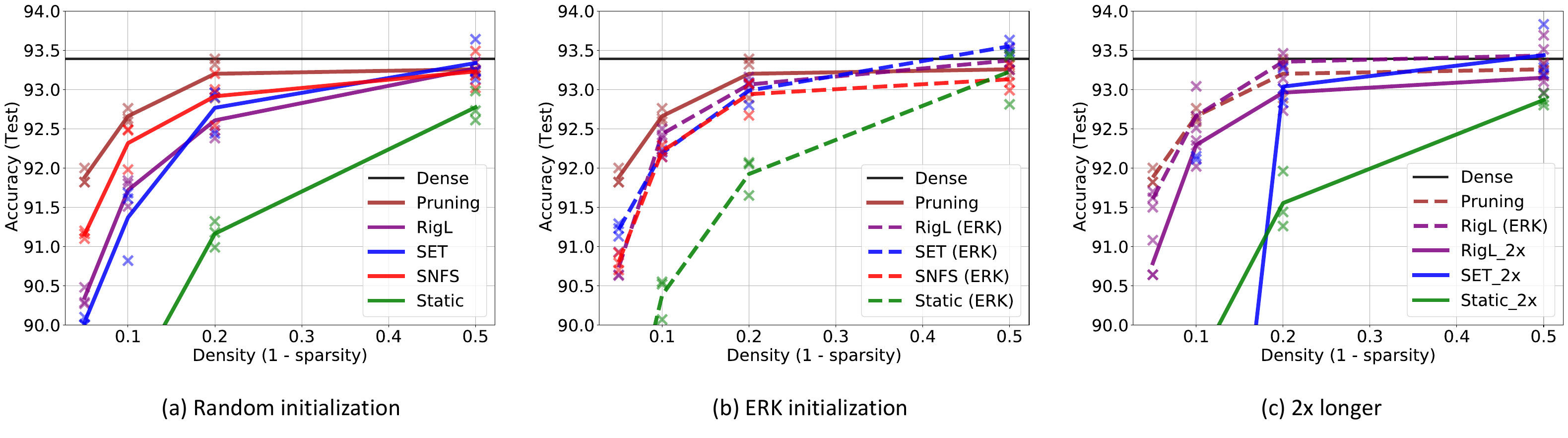}
    \captionsetup{aboveskip=\figureaboveskip,belowskip=\figurebelowskip}
    \caption{\textbf{Test Accuracy vs Sparsity on CIFAR-10,} plotted for Random initialization \textbf{(left)}, ERK initialization \textbf{(center)}, and for training $2\times$ longer \textbf{(right)}. Owing to random growth, SET can be unstable when training for longer durations with higher sparsities. Overall, \textit{RigL}\textsubscript{$2 \times$} (ERK) achieves highest test accuracy.}
    \label{fig:cifar10-main-results}
\end{figure}

%% file: sections/results_cifar100.tex
\subsection{ResNet-50 on CIFAR100}\label{cifar-100-results}

\begin{tabularx}{\textwidth}[hb]{*{2}{>{\centering\arraybackslash}X}}
    \centering
    \captionsetup{labelformat=andfigure,width=1.2\linewidth,aboveskip=7pt,belowskip=0pt}
    \captionlistentry[figure]{entry for figure}
    \label{fig:cifar100-main-results}
    
    \captionof{table}{\textbf{Benchmarking sparse ResNet-50s on CIFAR-100,} tabulated by performance and cost \textbf{(below)}, and plotted across densities \textbf{(right)}. In each group below, \textit{RigL} outperforms or matches existing sparse-to-sparse and dense-to-sparse methods. Notably, \textit{RigL}\textsubscript{$3\times$} at 90\% sparsity and \textit{RigL}\textsubscript{$2\times$} at 80\% sparsity surpass iterative pruning with similar FLOP consumption. \textit{RigL}\textsubscript{$2\times$} (ERK) further improves performance but requires a larger training budget. }
    \input{tables/cifar100}
    \label{tab:cifar100-main-results}  
&
    \input{figs/cifar100_main}
\end{tabularx}

We see similar trends when training sparse variants of ResNet-50 on the CIFAR-100 dataset (Table \ref{tab:cifar100-main-results}, metrics reported as in Section \ref{cifar-10-results}). We also include a comparison against sparse networks trained with the Lottery Ticket Hypothesis (\citet{frankle2018lottery}) in Table \ref{tab:cifar100-main-results}---we obtain tickets with a commensurate performance for sparsities lower than 80\%. Finally, the choice of initialization scheme affects the performance and FLOP consumption by a greater extent than the method used itself, with the exception of SNFS (groups 1 and 2 in Table \ref{tab:cifar100-main-results}). 

% Such networks are trained twice: once to prune the dense model, and then after re-initializing with the obtained mask. Finally, concurring with \citet{rigl}, multiple sparse methods generalize better than the dense baseline with just half the parameters, indicating the regularization aspect of sparse networks (Figure \ref{fig:cifar100-main-results}, bottom).  

% ERK initialization improves performance across all sparse methods, especially at higher sparsities. 

%% file: tables/cifar100.tex
\resizebox{1.15\linewidth}{!}{%
\begin{tabular}{ c cc cc }
\toprule
\multirow{3}{*}{\textbf{Method}}& 
\multicolumn{2}{c}{$\mathbf{1 - s=0.1}$} & \multicolumn{2}{c}{$\mathbf{1 - s=0.2}$} \\
\cmidrule(lr){2-3} \cmidrule(lr){4-5}
{} & 
\makecell{Accuracy $\uparrow$ \\ (Test)}  & \makecell{FLOPs $\downarrow$  \\ (Train, Test)} &
\makecell{Accuracy $\uparrow$ \\ (Test)}  & \makecell{FLOPs $\downarrow$  \\ (Train, Test)} \\
\midrule
Static & 
{69.7 $\pm$ 0.42} & {0.10x, 0.10x} & 
{72.3 $\pm$ 0.30} & {0.20x,0.20x} \\

Small Dense & 
{70.8 $\pm$ 0.22} & {0.11x, 0.11x} & 
{72.6$\pm$ 0.93} & {0.20x, 0.20x} \\

SET &
{71.4 $\pm$ 0.35} & {0.10x, 0.10x} & 
{73.4 $\pm$ 0.45} & {0.20x, 0.20x} \\

\textbf{RigL} &
\textbf{71.8 $\pm$ 0.33} & {0.10x, 0.10x} &
\textbf{73.5 $\pm$ 0.04} & {0.20x, 0.20x} \\
\midrule

Static (ERK) & 
{71.5 $\pm$ 0.18} & {0.22x, 0.22x} & 
{73.2 $\pm$ 0.39} & {0.38x, 0.38x} \\

SET (ERK)&
{72.3 $\pm$ 0.39} & {0.22x, 0.22x} &
\textbf{73.5 $\pm$ 0.25} & {0.38x, 0.38x} \\

\textbf{RigL (ERK)} &
\textbf{72.6 $\pm$ 0.37} & {0.23x, 0.22x} &
{73.4 $\pm$ 0.15} & {0.38x, 0.38x} \\
\midrule

SNFS & 
{72.3 $\pm$ 0.20} & {0.58x, 0.37x} & 
{73.9 $\pm$ 0.20} & {0.70x, 0.55x} \\ 

SNFS (ERK)& 
{73.0 $\pm$ 0.33} & {0.59x, 0.38x} & 
{73.9 $\pm$ 0.27} & {0.69x, 0.54x} \\

{Pruning} & 
{73.1 $\pm$ 0.32} & {0.36x,0.11x} & 
{73.8 $\pm$ 0.23} & {0.45x,0.25x} \\ 

{RigL\textsubscript{$2 \times$}} &
{73.1 $\pm$ 0.71} & {0.20x, 0.10x} &
{74.0 $\pm$ 0.24} & {0.41x, 0.20x} \\

{Lottery} &
{73.6 $\pm$ 0.32} & {0.62x,0.11x} & 
{74.2 $\pm$ 0.41} & {0.81x,0.25x} \\

\textbf{RigL\textsubscript{$3 \times$}} &
\textbf{73.7 $\pm$ 0.16} & {0.30x, 0.10x} &
{74.2 $\pm$ 0.23} & {0.61x, 0.20x} \\

\textbf{RigL\textsubscript{$2 \times$} (ERK)} &
{73.6 $\pm$ 0.05} & {0.46x, 0.22x} &
\textbf{74.4 $\pm$ 0.10} & {0.76x, 0.38x} \\

\midrule

\textbf{Dense Baseline} &
\textbf{74.7 $\pm$ 0.38} & {7.77e9, 2.59e9} &
\textbf{-} & {-} \\
\bottomrule

\end{tabular}%
}

%% file: figs/cifar100_main.tex
\centering
\includegraphics[width=0.66\linewidth,valign=t]{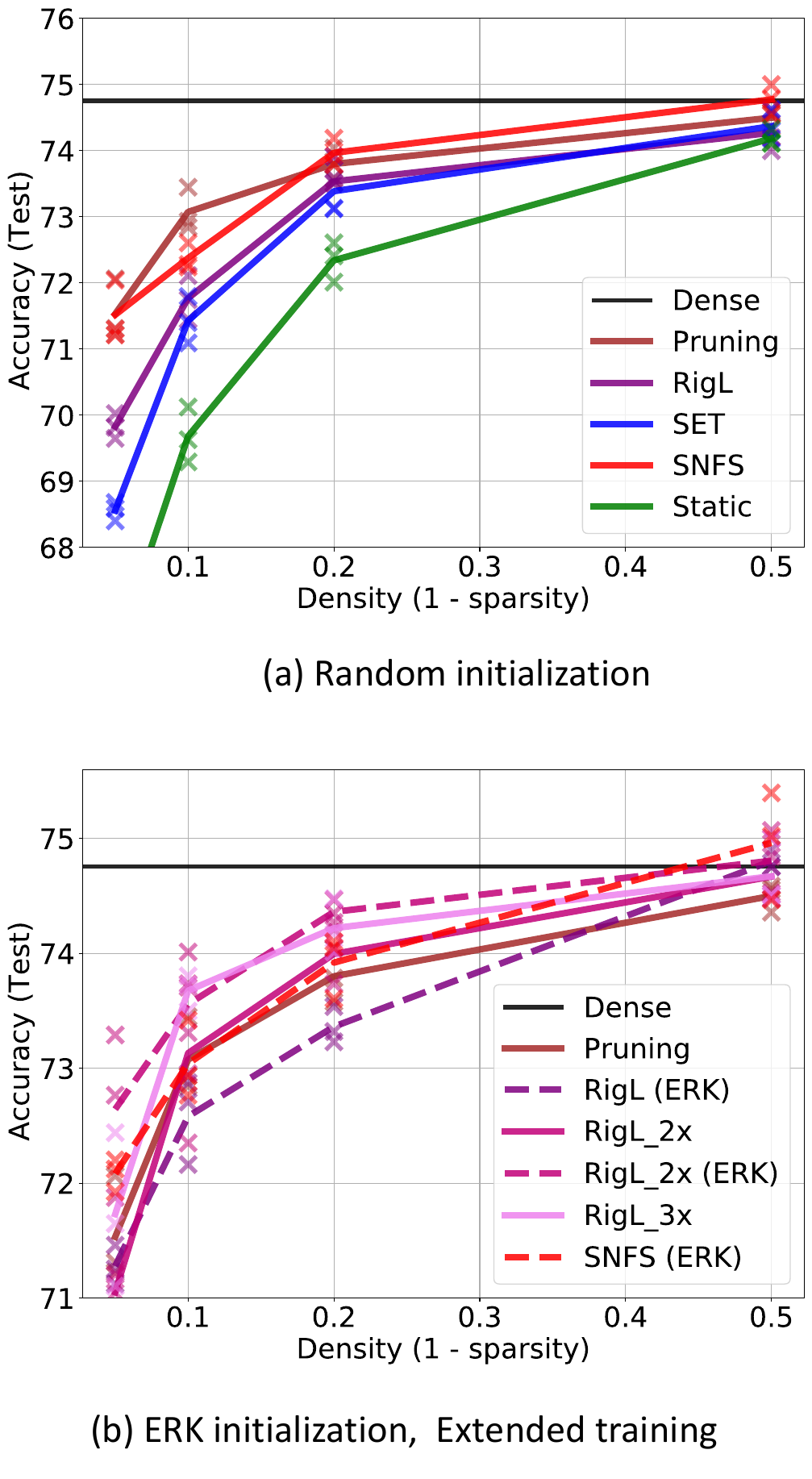}

%% file: sections/hyperparameter_tuning.tex
\subsection{Hyperparameter Tuning}\label{hyperparameter-tuning}

\input{tables/alpha_deltaT}

\paragraph{$(\alpha, \Delta T)$ vs Sparsities} To understand the impact of the two additional hyperparameters included in \textit{RigL}, we use a Tree of Parzen Estimator (TPE sampler, \citet{TPE_Bergstra}) via Optuna to tune $(\alpha, \Delta T)$. We do this for sparsities $(1 - s) \in \{0.1,0.2,0.5\}$, and a fixed learning rate of $0.1$. Additionally, we set the sampling domain for $\alpha$ and $\Delta T$ as $[0.1,0.6]$ and $\{50,100, 150,...,1000\}$ respectively. We use 15 trials for each sparsity value, with our objective function as the validation accuracy averaged across 3 random seeds.

\input{figs/lr_sweep}

Table \ref{tab:effect-alpha-deltaT} shows the test accuracies of tuned hyperparameters. While the reference hyperparameters (original authors, $\alpha=0.3, \Delta T=100$) differ from the obtained optimal hyperparameters, the difference in performance is marginal, especially for ERK initialization. This in agreement with the original paper, which finds $\alpha \in \{0.3, 0.5\}, \Delta T = 100$ to be suitable choices. We include contour plots detailing the hyperparameter trial space in the supplementary material.

\paragraph{Learning Rate vs Sparsities} We further examine if the final performance improves by tuning the learning rate ($\eta$) individually for each sparsity-initialization pair. We employ a grid search over $\eta \in \{0.1,0.05,0.01,0.005\}$ and $(\alpha, \Delta T) \in \{(0.3, 100), (0.4,200), (0.4, 500), (0.5, 750)\}$. As seen in Figure \ref{fig:lr-sweep}, $\eta = 0.1$ and $\eta = 0.05$ are close to optimal values for a wide range of sparsities and initializations. Since these learning rates also correspond to good choices for the Dense baseline, one can employ similar values when training with \textit{RigL}.

%% file: tables/alpha_deltaT.tex
\begin{table}[th]
    \captionsetup{aboveskip=\tableaboveskip,belowskip=\tablebelowskip}
    \caption{\textbf{Reference vs Optimal $(\alpha, \Delta T)$ on CIFAR-10.} Optimal hyperparameters are obtained by tuning with a TPE sampler in Optuna. The difference between the reference and optimal performance is small, indicating that there is not a significant benefit in tuning $(\alpha, \Delta T)$ individually for each initialization and sparsity configuration.}
    \label{tab:effect-alpha-deltaT}
    \centering
    
    \begin{tabular}{ c c  cc  cc}
    \toprule
    \multirow{3}{*}{\textbf{Initialization}} & \textbf{Density} & 
    \multicolumn{2}{c}{\textbf{Reference}} & \multicolumn{2}{c}{\textbf{Optimal}} \\
    \cmidrule(lr){2-2} \cmidrule(lr){3-4} \cmidrule(lr){5-6}
    {} & {$(1-s)$} & 
    {$(\alpha, \Delta T)$} & \makecell{Accuracy $\uparrow$ \\ (Test)} & 
    {$(\alpha, \Delta T)$} & \makecell{Accuracy $\uparrow$ \\ (Test)} \\
    \midrule
    
    Random & 0.1 & 
    {0.3, 100} & {91.7 $\pm$ 0.18} &  
    {0.197, 50} & \textbf{91.8 $\pm$ 0.17} \\
    
    Random & 0.2 & 
    {0.3, 100} & {92.6 $\pm$ 0.10} &  
    {0.448, 150} & \textbf{92.8 $\pm$ 0.16} \\
    
    Random & 0.5 & 
    {0.3, 100} & \textbf{93.3 $\pm$ 0.07} &  
    {0.459, 550} & \textbf{93.3 $\pm$ 0.18} \\
    \midrule
    
    ERK & 0.1 & 
    {0.3, 100} & \textbf{92.4 $\pm$ 0.06} &  
    {0.416, 200} & \textbf{92.4 $\pm$ 0.23} \\
    
    ERK & 0.2 & 
    {0.3, 100} & \textbf{93.1 $\pm$ 0.09} &  
    {0.381, 950} & \textbf{93.1 $\pm$ 0.21} \\
    
    ERK & 0.5 & 
    {0.3, 100} & {93.4 $\pm$ 0.14} &  
    {0.287, 500} & \textbf{93.8 $\pm$ 0.06} \\
    \hline

    \end{tabular}
    
    \label{tab:replication_verify}
\end{table}

%% file: figs/lr_sweep.tex
\begin{figure}[!b]
    \centering
    \includegraphics[width=\textwidth]{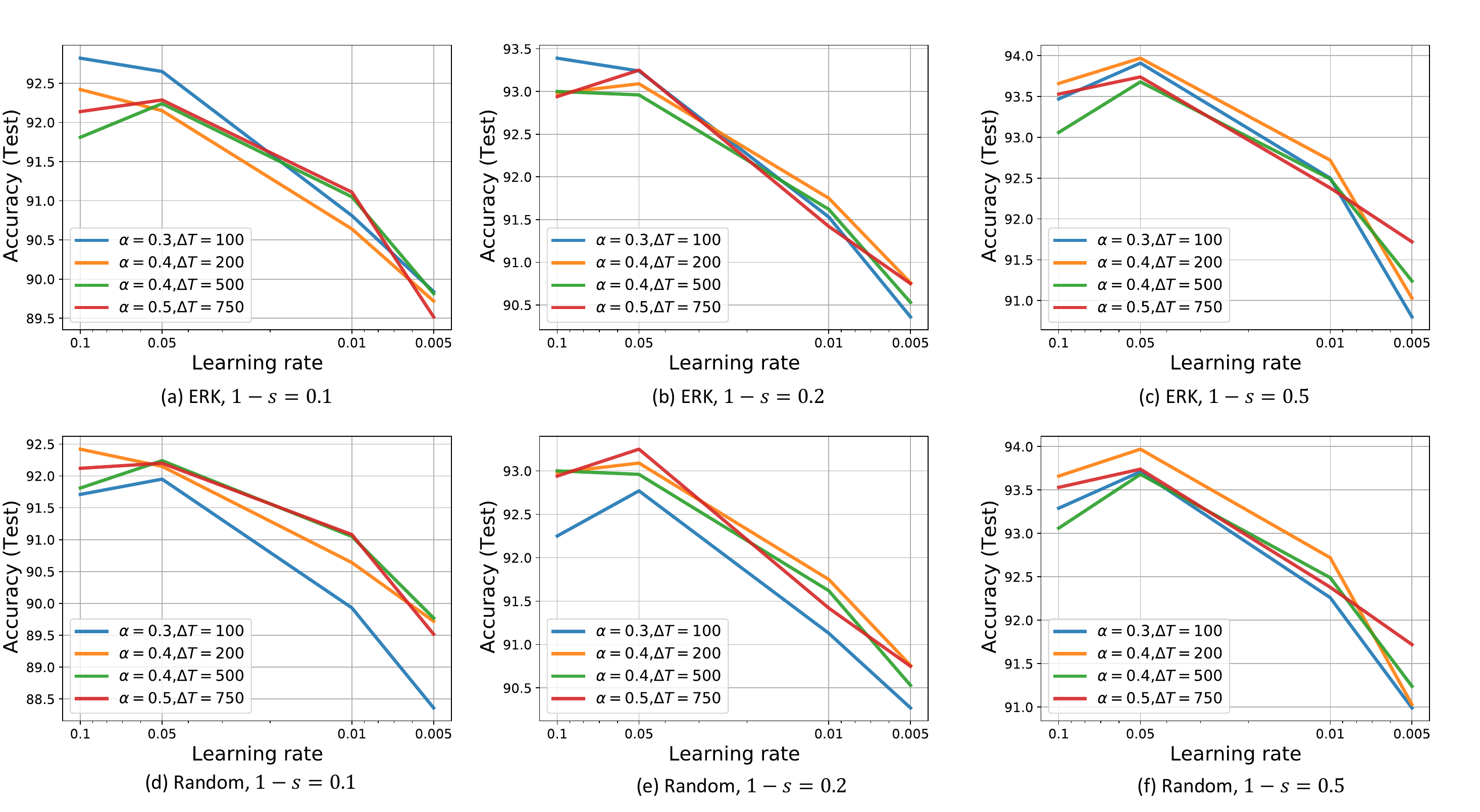}
    \captionsetup{aboveskip=\figureaboveskip,belowskip=\figurebelowskip}
    \caption{\textbf{Learning Rate vs Sparsity on CIFAR-10.} Runs using a learning rate $> 0.1$ do not converge and are not plotted here. There is little benefit in tuning the learning rate for each sparsity, and $0.1, 0.05$ are good choices overall.}
    \label{fig:lr-sweep}
\end{figure}

%% file: sections/results_beyond.tex
\section{Results beyond Original Paper}
\jdcomment{Often papers don't include enough information to fully specify their experiments, so some additional experimentation may be necessary. For example, it might be the case that batch size was not specified, and so different batch sizes should be evaluated. Include the results of any additional experiments here. Note: this won't be necessary for all reproductions.}
 
\subsection{Sparsity Distribution vs FLOP Consumption}\label{effect-sparsity-distribution}

\input{figs/erk_vs_random_FLOPs}

While ERK initialization outperforms Random initialization consistently for a given target parameter count, it requires a higher FLOP budget. Figure \ref{fig:erk-vs-random-FLOPs} compares the two initialization schemes across fixed training FLOPs. Theoretical FLOP requirement for Random initialization scales linearly with density $(1-s)$, and is significantly lesser than ERK's FLOP requirements. Consequently, Random initialization outperforms ERK initialization for a given training budget.

\subsection{Effect of Redistribution}\label{effect-redistribution}

\input{tables/effect_redistribution}

One of the main differences of \textit{RigL} over SNFS is the lack of layer-wise redistribution during training. We examine if using a redistribution criterion can be beneficial and bridge the performance gap between Random and ERK initialization. Following \cite{dettmers2020sparse}, during every mask update, we reallocate layer-wise density proportional to its average sparse gradient or momentum (\textit{RigL}-SG, \textit{RigL}-SM).
 
Table \ref{tab:effect-redistribution} shows that redistribution significantly improves \textit{RigL} (Random), but not \textit{RigL} (ERK). We additionally plot the FLOP requirement against training steps and the final sparsity distribution in Figure \ref{fig:density-dist-evolution}. The layer-wise sparsity distribution largely becomes constant within a few epochs. The final distribution is similar, but more ``extreme'' than ERK---wherever ERK exceeds/falls short of Random, redistribution does so by a greater extent.

% Like ERK, redistribution allocates higher densities for $1 \times 1$ convolutions (\textit{convShortcut} in Figure \ref{fig:density-dist-evolution}), resulting in increased FLOP consumption. 

% Figure \ref{fig:density-dist-evolution} illustrates this for 80\% sparse \textit{RigL}-SG and \textit{RigL}-SM on CIFAR-100 (both randomly initialized). 

By allocating higher densities to $1 \times 1$ convolutions (\textit{convShortcut} in Figure \ref{fig:density-dist-evolution}), redistribution significantly increases the FLOP requirement---and hence, is not a preferred alternative to ERK. Surprisingly, initializing \textit{RigL} with the final sparsity distribution in a manner similar to the Lottery Ticket Hypothesis results in subpar performance (group 3, Table \ref{tab:effect-redistribution}). 

\input{figs/density_dist_evolution}

%% file: figs/erk_vs_random_FLOPs.tex
\begin{figure}[h]
    \centering
    \includegraphics[width=1\textwidth]{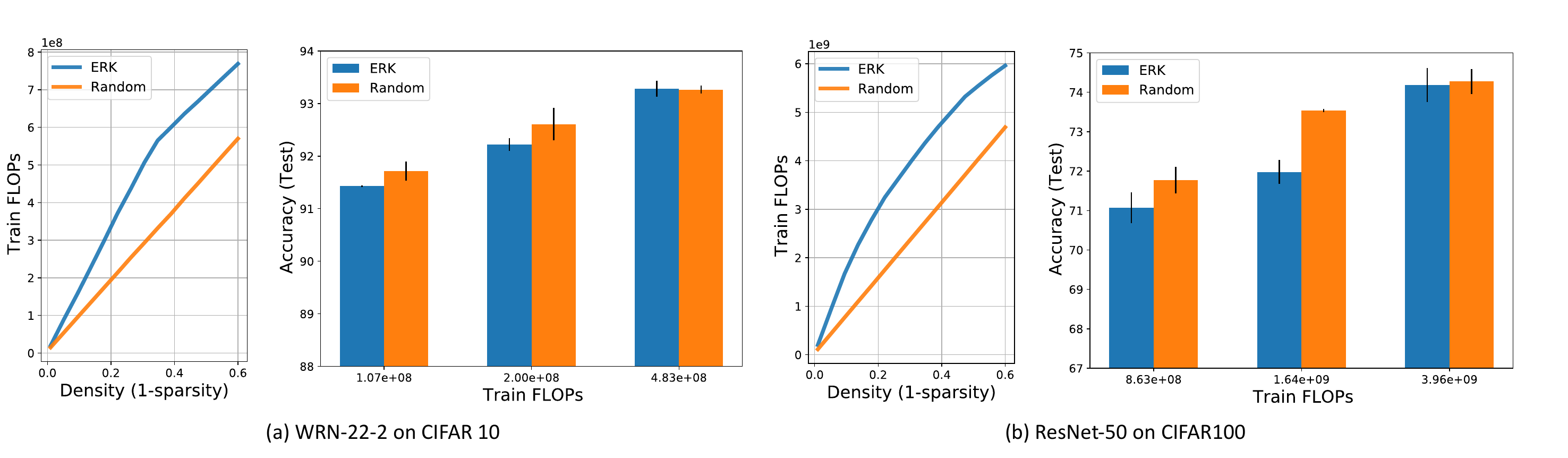}
    \captionsetup{aboveskip=\figureaboveskip,belowskip=\figurebelowskip}
    \caption{\textbf{Test Accuracy vs FLOP consumption of WideResNet-22-2 on CIFAR-10 and ResNet-50 on CIFAR-100,} compared for Random and ERK initializations. For the same FLOP budget, models trained with ERK initialization must be more sparse, resulting in inferior performance.}
    \label{fig:erk-vs-random-FLOPs}
\end{figure}

%% file: tables/effect_redistribution.tex
\begin{table}[h]
    \captionsetup{aboveskip=\tableaboveskip,belowskip=\tablebelowskip}
    \caption{\textbf{Effect of redistribution during \textit{RigL} updates, evaluated  on CIFAR10 and CIFAR100}. By utilising sparse gradient or sparse momentum based redistribution, \textit{RigL} (Random) matches \textit{RigL} (ERK)'s performance. Among Random and ERK initialized experiments, we mark the best metrics under each sparsity and dataset in bold. }
    \centering
    
    \resizebox{\textwidth}{!}{%
    \begin{tabular}{ c c cc cc cc cc}
     \toprule
    \multirow{3}{*}{\textbf{Method}} & \multirow{3}{*}{\textbf{Redistribution}} & 
    \multicolumn{4}{c}{\textbf{CIFAR-10}} & \multicolumn{4}{c}{\textbf{CIFAR-100}} \\
    \cmidrule(lr){3-6} \cmidrule(lr){7-10}
    
    {} & {} &
    \multicolumn{2}{c}{$1 - s=0.1$} & \multicolumn{2}{c}{$1 - s=0.2$} &
    \multicolumn{2}{c}{$1 - s=0.1$} & \multicolumn{2}{c}{$1 - s=0.2$} \\
    \cmidrule(lr){3-4} \cmidrule(lr){5-6} \cmidrule(lr){7-8} \cmidrule(lr){9-10}
    
    {} & {} &
    \makecell{Accuracy $\uparrow$ \\ (Test)}  & \makecell{FLOPs $\downarrow$  \\ (Train, Test)} &
    \makecell{Accuracy $\uparrow$ \\ (Test)}  & \makecell{FLOPs $\downarrow$  \\ (Train, Test)}  &
    \makecell{Accuracy $\uparrow$ \\ (Test)}  & \makecell{FLOPs $\downarrow$  \\ (Train, Test)} &
    \makecell{Accuracy $\uparrow$ \\ (Test)}  & \makecell{FLOPs $\downarrow$  \\ (Train, Test)} \\
    \midrule
    
    \multicolumn{10}{c}{Random Initialization} \\
    \midrule
    
    {RigL} & {-} &
    {91.7 $\pm$ 0.18} & {0.10x, 0.10x} &
    {92.9 $\pm$ 0.10} & {0.20x, 0.20x} &
    {71.8 $\pm$ 0.33} & {0.10x, 0.10x} &
    {73.5 $\pm$ 0.04} & {0.20x, 0.20x} \\
    
    {RigL-SG} & {Sparse Grad} &
    \textbf{92.2 $\pm$ 0.17} & {0.28x, 0.28x} &
    {92.7 $\pm$ 0.25} & {0.49x, 0.49x} &
    {72.3 $\pm$ 0.12} & {0.36x,0.35x} &
    \textbf{73.7 $\pm$ 0.15} & {0.53x, 0.53x} \\
    
    {RigL-SM} & {Sparse Mmt} &
    \textbf{92.2 $\pm$ 0.20} & {0.28x, 0.28x} &
    \textbf{92.9 $\pm$ 0.21} & {0.50x, 0.49x} &
    \textbf{72.6 $\pm$ 0.27} & {0.36x,0.36x} &
    \textbf{73.7 $\pm$ 0.35} & {0.53x, 0.53x} \\
    
    \midrule
    \multicolumn{10}{c}{ERK Initialization} \\
    \midrule
    
    {RigL} & {-} &
    \textbf{92.4 $\pm$ 0.06} & {0.17x, 0.17x} &
    \textbf{93.1 $\pm$ 0.09} & {0.35x, 0.35x} &
    {72.6 $\pm$ 0.37} & {0.23x, 0.22x} &
    {73.4 $\pm$ 0.15} & {0.38x, 0.38x} \\
    
    {RigL-SG} & {Sparse Grad} &
    {92.1 $\pm$ 0.19} & {0.28x, 0.28x} &
    {92.7 $\pm$ 0.19} & {0.49x, 0.49x} &
    \textbf{73.0 $\pm$ 0.13} & {0.37x,0.36x} &
    {74.2 $\pm$ 0.26} & {0.53x, 0.53x} \\
    
    {RigL-SM} & {Sparse Mmt} &
    \textbf{92.27 $\pm$ 0.01} & {0.28x, 0.28x} &
    \textbf{93.0 $\pm$ 0.13} & {0.50x, 0.49x} &
    {72.6 $\pm$ 0.27} & {0.37x, 0.37x} &
    \textbf{74.2 $\pm$ 0.13} & {0.53x, 0.53x} \\
    
    \midrule
    \multicolumn{10}{c}{Re-Initialization with \textit{RigL}-SM (Random, ERK)} \\
    \midrule
    
    \makecell{RigL} & {-} &
    {90.3 $\pm$ 0.34} & {0.28x, 0.28x} &
    {91.0 $\pm$ 0.38} & {0.50x, 0.49x} &
    {67.6 $\pm$ 0.28} & {0.36x, 0.36x} &
    {68.9 $\pm$ 0.65} & {0.53x, 0.53x} \\
    
    \makecell{RigL (ERK)} & {-} &
    {90.2 $\pm$ 0.57} & {0.28x, 0.28x} &
    {90.6 $\pm$ 0.56} & {0.50x, 0.49x} &
    {67.8 $\pm$ 0.73} & {0.37x, 0.37x} &
    {68.9 $\pm$ 0.47} & {0.53x, 0.53x} \\
    \bottomrule
    
    \end{tabular}%
    }
    
    \label{tab:effect-redistribution}
\end{table}

%% file: figs/density_dist_evolution.tex
\begin{figure}[!ht]
    \centering
    \includegraphics[width=1\textwidth]{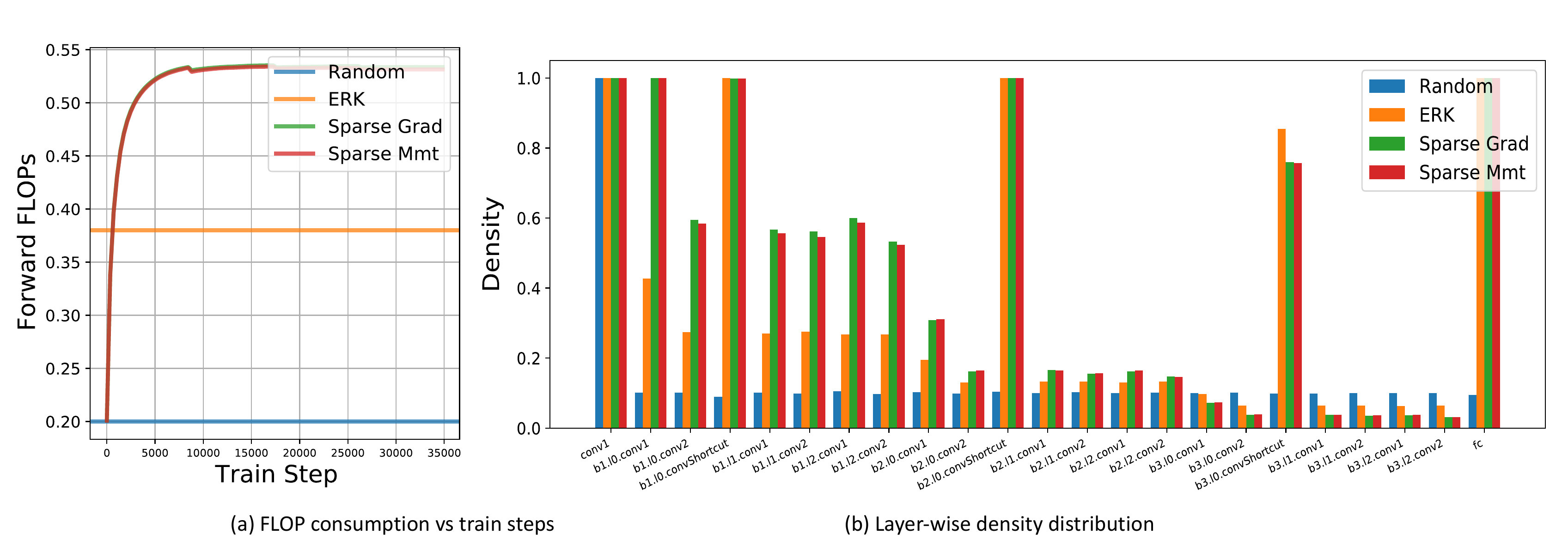}
    \captionsetup{aboveskip=\figureaboveskip,belowskip=\figurebelowskip}
    \caption{\textbf{Effect of redistribution on \textit{RigL}'s performance,} evaluated using WideResNet-22-2 on CIFAR10 at 80\% sparsity. \textbf{(left)} FLOPs required per forward pass, shown relative to the dense baseline, rises quickly and saturates within a few epochs (\textasciitilde10k steps) for both sparse gradient and sparse momentum based redistribution. \textbf{(right)} Comparison of the final density distribution against Random and ERK counterparts. ``b'' refers to block and ``l'' layer here.}
    \label{fig:density-dist-evolution}
\end{figure}

%% file: sections/discussion.tex
\section{Discussion}

\vscomment{Conclusions should synthesize the results of your paper and separate what is significant from what is not. Ideally, they should add new information and observations that put your results in perspective. Here's a simple test: if somebody reads your conclusions before reading the rest of your paper, will they fully understand them? If the answer is ``yes,'' there's probably something wrong. A good conclusion says things that become significant after the paper has been read. A good conclusion gives perspective to sights that haven't yet been seen at the introduction. A conclusion is about the implications of what the reader has learned. Of course, a conclusion is also an excellent place for conjectures, wish lists, and open problems.}

\jdcomment{Give your judgement on if your experimental results support the claims of the paper. Discuss the strengths and weaknesses of your approach - perhaps you didn't have time to run all the experiments, or perhaps you did additional experiments that further strengthened the claims in the paper.}

Evaluated on image classification, the central claims of \citet{rigl} hold true---\textit{RigL} outperforms existing sparse-to-sparse training methods and can also surpass other dense-to-sparse training methods with extended training. \textit{RigL} is fairly robust to its choice of hyperparameters, as they can be set independent of sparsity or initialization. We find that the choice of initialization has a greater impact on the final performance and compute requirement than the method itself. Considering the performance boost obtained by redistribution, proposing distributions that attain maximum performance given a FLOP budget could be an interesting future direction.

% Finally, our early experiments show that adapting \textit{RigL} to promote structured sparsity can obtain practical speedups in existing hardware accelerators. 

For computational reasons, our scope is restricted to small datasets such as CIFAR-10/100. \textit{RigL}'s applicability outside image classification---in Computer Vision and beyond (machine translation etc.) is not covered here.

\paragraph{What was easy}
\jdcomment{Give your judgement of what was easy to reproduce. Perhaps the author's code is clearly written and easy to run, so it was easy to verify the majority of original claims. Or, the explanation in the paper was really easy to follow and put into code. 

Be careful not to give sweeping generalizations. Something that is easy for you might be difficult to others. Put what was easy in context and explain why it was easy (e.g. code had extensive API documentation and a lot of examples that matched experiments in papers). }

The authors' code covered most of the experiments in their paper and helped us validate the correctness of our replicated codebase. Additionally, the original paper is quite complete, straightforward to follow, and lacked any major errors.

\paragraph{What was difficult}
\jdcomment{List part of the reproduction study that took more time than you anticipated or you felt were difficult. 

Be careful to put your discussion in context. For example, don't say "the maths was difficult to follow", say "the math requires advanced knowledge of calculus to follow".}

Implementation details such as whether momentum buffers were accumulated sparsely or densely had a substantial impact on the performance of SNFS. Finding the right $\epsilon$ for ERK initialization required handling of edge cases---when a layer's capacity is exceeded. Hyperparameter tuning $(\alpha, \Delta T)$ involved multiple seeds and was compute-intensive.

% \paragraph{Discrepancies} SNFS redistributes layer-wise density during training, and hence its inference FLOPs do not scale proportional to density, i.e.,  $f_s \neq f_d * (1 - s)$. Similarly, Pruning inference FLOPs do not scale with density.

\paragraph{Communication with original authors}

\jdcomment{Document the extent of (or lack of) communication with the original authors. To make sure the reproducibility report is a fair assessment of the original research we recommend getting in touch with the original authors. You can ask authors specific questions, or if you don't have any questions you can send them the full report to get their feedback before it gets published. }

We acknowledge and thank the original authors for their responsive communication, which helped clarify a great deal of implementation and evaluation specifics. Particularly, FLOP counting for various methods while taking into account the changing sparsity distribution. We also discussed experiments extending the original paper---as to whether the authors had carried out a similar study before.

%% file: supplementary_sections/architecture_details.tex
\section{Architecture Specific Details---ResNet-50 on CIFAR100}

\begin{table}[ht]
\captionsetup{aboveskip=\tableaboveskip,belowskip=\tablebelowskip}
\caption{\textbf{ResNet-50 architecture used on CIFAR100}. Building blocks are shown in brackets, with the numbers of blocks stacked. Downsampling is performed by conv3\_1, conv4\_1, and conv5\_1 with a stride of 2.
}
\label{tab:architecture}
\centering
\begin{tabular}{c c c}
    \toprule
    Layer Name & Output Size & ResNet-50 \\
    \midrule
    
    conv1 & 32$\times$32 & {3$\times$3, 64, no stride}\\
    \midrule
    
    \multirow{3}{*}{conv2\_x} & \multirow{3}{*}{32$\times$32} 
    & \blockb{256}{64}{3} \\
    &  &  \\
    &  &  \\
    \midrule
    
    \multirow{3}{*}{conv3\_x} &  \multirow{3}{*}{16$\times$16}  
    & \blockb{512}{128}{4} \\
    &  &  \\
    &  &  \\
    \midrule
    
    \multirow{3}{*}{conv4\_x} & \multirow{3}{*}{8$\times$8}  
    & \blockb{1024}{256}{6} \\
      &  &  \\
      &  &  \\
    \midrule
    
    \multirow{3}{*}{conv5\_x} & \multirow{3}{*}{4$\times$4}  
    & \blockb{2048}{512}{3} \\
      &  &   \\
      &  &   \\
    \midrule
    
    & 1$\times$1  & {average pool, 100-d fc, softmax} \\
    \midrule
    
    \multicolumn{2}{c}{FLOPs} & 2.59e9 \\
    \bottomrule
\end{tabular}
\end{table}

We use a variant of the originally proposed ResNet architecture (\citet{He_2016_CVPR}). Particularly, we replace the initial $7 \times 7$ conv layer with a $3 \times 3$ conv layer. Here, ``conv layer'' refers to convolution followed by batchnorm (\citet{ioffe2015batch}) and ReLU activation. This is intended to not excessively downsample the image---CIFAR-100 (\citet{Krizhevsky09learningmultiple}) has images of dimensions $32 \times 32$, compared to Imagenet's (\citet{ILSVRC15}) $224 \times 224$. Each block used (conv2\_x, conv3\_x, etc.) is a bottleneck block, and uses the conv-batchnorm-ReLU ordering. 

%% file: supplementary_sections/flop_counting.tex
\section{FLOP Counting Procedure}

Following \citet{rigl}, we base our counting procedure on the Micronet Challenge\footnote{https://micronet-challenge.github.io/}, which was conducted as a part of NeurIPS 2019. Support for unstructured sparsity is assumed while computing the number of additions and multiplication operations. The sum of these two gives us the theoretical FLOPs for a single forward pass through the model. 

Concretely, let the FLOPs required for a forward pass through a dense model be $f_d$ and the corresponding for a sparse model (or small-dense model) be $f_s$. Then, the FLOPs for training a dense model are $3f_d$---since the backward pass involves computing gradients with respect to each weight and activation. $f_s$ can be computed for a model given its sparsity distribution via the counting procedure. The FLOPs required to train a sparse model depend on the technique used, as detailed below.

% Below, we detail the procedure for computing inference and train FLOPs of specific methods.

\subsection{Inference FLOPs}

% As mentioned above, we determine the inference FLOPs of each model given its layer-wise sparsity distribution (specified by the boolean masks). Below we highlight the 

\paragraph{Small-Dense, RigL, SET, Static} These methods involve constant layer-wise sparsity throughout training, hence the FLOP count can be determined during any step. The FLOP count for Random initialized models are $(1-s)$ times the Dense FLOPs.

\paragraph{SNFS, Pruning} Both methods involve varying layer-wise sparsity during training, and hence non-constant FLOP consumption. The final weights are used to determine inference FLOPs in this case.

\subsection{Train FLOPs}

\paragraph{Small-Dense, Static} Dense gradients are not required by these models, and hence have a train FLOP count of $3f_s$.

\paragraph{SET} Dense gradients are not required, and random growth can be implemented quite efficiently. Thus, the train FLOP count is $3f_s$.

\paragraph{RigL} Dense gradients are required only every $\Delta T$ steps, hence the corresponding train FLOP count is: $\frac{3\Delta Tf_s + 2f_s + f_d }{\Delta T + 1}$. We note that since $\Delta T$ is typically set between 100--1000, the preceding expression is quite close to $3f_s$.

\paragraph{SNFS} Dense gradients are required at each training step, resulting in $2f_s + f_d$ FLOPs consumed at each step. Since the sparse FLOP count varies as we train, the average FLOP count is: $2\mathbb{E}[f_{s,t}] + f_d$, where $f_{s,t}$ is the sparse inference FLOPs at train step $t$.

\paragraph{Pruning} Does not require dense gradients, but the sparsity increases smoothly from $0\%$ to the target value as we train. The FLOP consumption here is $3\mathbb{E}[f_{s,t}]$, , where $f_{s,t}$ is the sparse inference FLOPs at train step $t$.

To determine $\mathbb{E}[f_{s,t}]$, we compute a running average of the FLOP consumption after every epoch. Notably, we find that the inference cost of Pruning is often close to a Random initialized sparse network, while SNFS, regardless of initiazation, is compute-intensive.

%% file: supplementary_sections/hyperparameter_tuning.tex
\section{Trial Space of Hyperparameter Tuning}

\begin{figure}[!h]
    \centering
    \includegraphics[width=\textwidth]{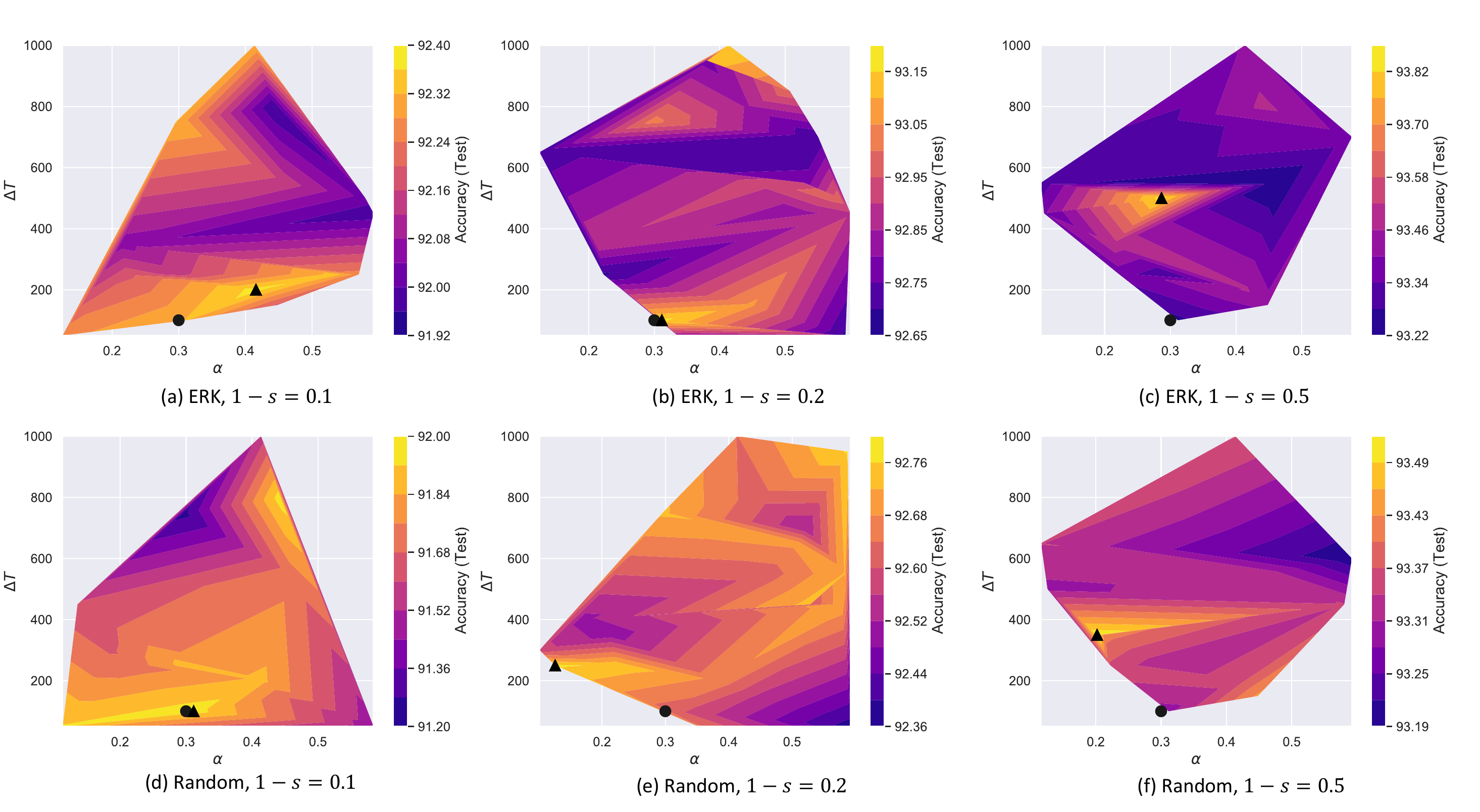}
    \captionsetup{aboveskip=\figureaboveskip,belowskip=\figurebelowskip}
    \caption{\textbf{Trial space of tuning $(\alpha, \Delta T)$}, shown as a countor plot. Here, black circle corresponds to $(\alpha=0.3,\Delta T = 100)$, while black triangle corresponds to the optimal hyper-parameter pair found. We plot the convex hull of the trial space, so in a few cases the reference point lies on the border of this space.}
    \label{fig:alpha-deltaT-contour}
\end{figure}

Figure \ref{fig:alpha-deltaT-contour} shows the hyper-parameter study for tuning $(\alpha, \Delta T)$ as a contour plot. We observe that for multiple initialization-density configurations, the reference choice $(\alpha=0.3,\Delta T = 100)$, is quite close to the optimal hyper-parameters. Furthermore, where they differ, the difference is within standard deviation bounds (Table 4 of the main report).

%% file: supplementary_sections/structured_sparsity.tex
\section{Dynamic Structured Sparsity}\label{structured-sparsity}

\input{supplementary_sections/table_structured_sparsity}

Present hardware accelerators lack efficient implementations for unstructured sparsity. As a result, in practice, the reduced FLOP requirement of sparse methods rarely translate to wall-clock improvements. In comparison, there are efficient implementations available for structured (or block) sparsity which reach theoretical speedups (\citet{gray2017gpu,Vooturi_2019_ICCV}). Motivated by this, we try modifying \textit{RigL} to explicitly work on structured sparsity. We promote channel sparsity for convolutional layers and keep fully connected layers dense. Mask update steps also operate at the channel level, based on \textit{RigL}'s growth and pruning criterion. We name this method as \textit{RigL}-struct. Such an approach is enticing, as we can remove masked-out channels, and obtain practical speedups on accelerators without needing support for unstructured sparsity.

Unfortunately, \textit{RigL}-struct does not preserve the performance of originally proposed \textit{RigL} (Table \ref{tab:structured-sparsity}). In fact, it performs only as good as Small-Dense models, which negates the motivation behind such an experiment---Small-Dense models already achieve the intended speedups.

%% file: supplementary_sections/table_structured_sparsity.tex
\begin{table}[h]
    \captionsetup{aboveskip=\tableaboveskip,belowskip=\tablebelowskip}
    \caption{\textbf{Modifying \textit{RigL} for structured sparsity, compared on CIFAR-10 and CIFAR-100 datasets.} \textit{RigL}-struct fails to match the accuracy of \textit{RigL} and just matches Small-Dense in performance. }
    \centering
    
    \resizebox{\textwidth}{!}{%
    \begin{tabular}{ c cc cc cc cc}
     \toprule
    \multirow{3}{*}{\textbf{Method}} & 
    \multicolumn{4}{c}{\textbf{CIFAR-10}} & \multicolumn{4}{c}{\textbf{CIFAR-100}} \\
    \cmidrule(lr){2-5} \cmidrule(lr){6-9}
    
    {} &
    \multicolumn{2}{c}{$1 - s=0.1$} & \multicolumn{2}{c}{$1 - s=0.2$} &
    \multicolumn{2}{c}{$1 - s=0.1$} & \multicolumn{2}{c}{$1 - s=0.2$} \\
    \cmidrule(lr){2-5} \cmidrule(lr){4-5} \cmidrule(lr){6-7} \cmidrule(lr){8-9}
    
    {} & 
    \makecell{Accuracy $\uparrow$ \\ (Test)}  & \makecell{Wall Time $\downarrow$} &
    \makecell{Accuracy $\uparrow$ \\ (Test)}  & \makecell{Wall Time $\downarrow$}  &
    \makecell{Accuracy $\uparrow$ \\ (Test)}  & \makecell{Wall Time $\downarrow$} &
    \makecell{Accuracy $\uparrow$ \\ (Test)}  & \makecell{Wall Time $\downarrow$} \\
    \midrule
    
    {Small-Dense} &
    {89.0 $\pm$ 0.35} & {0.11x} & 
    {91.0 $\pm$ 0.07} & {0.20x} &
    {70.8 $\pm$ 0.22} & {0.11x} & 
    {72.6$\pm$ 0.93} & {0.20x} \\
    
    \midrule
    
    \multicolumn{9}{c}{Random Initialization} \\
    \midrule
    
    {RigL} &
    {91.7 $\pm$ 0.18} & {1.0x} &
    {92.9 $\pm$ 0.10} & {1.0x} &
    {71.8 $\pm$ 0.33} & {1.0x} &
    {73.5 $\pm$ 0.04} & {1.0x} \\
    
    {RigL-Struct} &
    {87.0 $\pm$ 0.09} & {0.10x} &
    {90.4 $\pm$ 0.27} & {0.20x} &
    {69.1 $\pm$ 0.11} & {0.10x} &
    {71.9 $\pm$ 0.13} & {0.20x} \\
    
    \midrule
    \multicolumn{9}{c}{ERK Initialization} \\
    \midrule
    
    {RigL} &
    {92.4 $\pm$ 0.06} & {1.0x} &
    {93.1 $\pm$ 0.09} & {1.0x} &
    {72.6 $\pm$ 0.37} & {1.0x} &
    {73.4 $\pm$ 0.15} & {1.0x} \\
    
    {RigL-Struct} &
    {89.6 $\pm$ 0.16} & {0.17x} &
    {91.3 $\pm$ 0.18} & {0.35x} &
    {71.1 $\pm$ 0.15} & {0.23x} &
    {72.9 $\pm$ 0.08} & {0.38x} \\
    \bottomrule
    
    \end{tabular}%
    }
    
    \label{tab:structured-sparsity}
\end{table}